\documentclass[letterpaper]{article} 
\usepackage[preprint]{aaai2027}  
\usepackage[hyphens]{url}  
\usepackage{graphicx} 
\usepackage{natbib}  
\usepackage{caption} 
\usepackage{amsmath,amssymb,amsfonts}
\usepackage{booktabs}
\usepackage{array}
\usepackage[table]{xcolor}
\usepackage{textcomp}
\usepackage{multirow}
\newcommand{\Ldet}[1]{\mathcal{L}_{\mathrm{det}}^{#1}}
\newcommand{\Lmap}[1]{\mathcal{L}_{\mathrm{map}}^{#1}}
\newcommand{\Lvlm}{\mathcal{L}_{\mathrm{VLM}}}
\newcommand{\Lplan}{\mathcal{L}_{\mathrm{plan}}}

\newcolumntype{G}{>{\columncolor{gray!12}}c}

\newcommand{\cmark}{\checkmark}
\newcommand{\xmark}{$\times$}

\title{Roadside-Cooperative Autonomous Driving: From Data Platform to Vision-Language End-to-End Reasoning}
\author{
    Yitao Xu\textsuperscript{\rm 1},
    Tong Wu\textsuperscript{\rm 1},
    Yiyan Wu\textsuperscript{\rm 1},
    Guoji Xu\textsuperscript{\rm 2},
    Yanbo Jiang\textsuperscript{\rm 1},
    Jiahao Wang\textsuperscript{\rm 1},\\
    Zehong Ke\textsuperscript{\rm 1},
    Junkai Jiang\textsuperscript{\rm 1},
    Fang Zhang\textsuperscript{\rm 3}\corresponding,
    Jianqiang Wang\textsuperscript{\rm 1}\corresponding\\
}
\affiliations{
    \textsuperscript{\rm 1}School of Vehicle and Mobility, Tsinghua University\\
    \textsuperscript{\rm 2}Suzhou Automotive Research Institute, Tsinghua University\\
    \textsuperscript{\rm 3}State Key Laboratory of Intelligent Vehicle and Mobility, Tsinghua University\\
    xuyt21@mails.tsinghua.edu.cn, zhang\_fang@tsinghua.edu.cn, wjqlws@tsinghua.edu.cn
}

\begin{document}
\maketitle

\begin{abstract}
Vehicle-to-Everything (V2X) cooperation enables beyond-line-of-sight perception, mitigating occlusions in single-vehicle sensing. However, existing V2X benchmarks provide limited support for closed-loop evaluation and language-grounded supervision, hindering the development of vision-language models (VLMs) for end-to-end cooperative driving. To address these limitations, we introduce V2XBench, a simulation platform featuring synchronized ego--roadside sensing and closed-loop evaluation, together with Chat-V2XBench, a progressively structured VQA dataset for cooperative reasoning. Building upon this benchmark infrastructure, we propose AURORA, an end-to-end cooperative driving framework. Equipped with a dual-view perception architecture, AURORA mitigates spatial and semantic discrepancies across ego and roadside viewpoints through a query-level Cross-View Query Alignment and Fusion (CQAF) module. Leveraging the resulting unified tokens, a LoRA-adapted VLM bridges semantic reasoning and generative trajectory planning. Extensive closed-loop evaluations on V2XBench demonstrate that AURORA achieves state-of-the-art performance in heavily occluded scenarios, with a Route Completion rate of 98.21\% and a Driving Score of 76.02, while requiring low roadside communication bandwidth. Ultimately, this work pioneers an extensible V2X--VLM paradigm, paving the way for next-generation cooperative autonomous driving.
\end{abstract}

\begin{table*}[t]
\centering
\caption{Comparison with representative public autonomous driving and V2X datasets. V2XBench is the only V2X benchmark to evaluate closed-loop E2E simulation with VQA capabilities.}
\label{tab:v2xbench-comparison}
\small
\begin{tabular*}{\textwidth}{@{\extracolsep{\fill}}lccccccccc@{}}
\hline
\textbf{Benchmark} &
\textbf{Year} &
\begin{tabular}[c]{@{}c@{}}\textbf{Real/}\\\textbf{Sim.}\end{tabular} &
\textbf{View} &
\begin{tabular}[c]{@{}c@{}}\textbf{Scenario}\\\textbf{Types}\end{tabular} &
\begin{tabular}[c]{@{}c@{}}\textbf{Closed-}\\\textbf{Loop}\end{tabular} &
\begin{tabular}[c]{@{}c@{}}\textbf{E2E-}\\\textbf{Sim}\end{tabular} &
\begin{tabular}[c]{@{}c@{}}\textbf{With}\\\textbf{Traffic Light}\end{tabular} &
\begin{tabular}[c]{@{}c@{}}\textbf{With}\\\textbf{VRUs}\end{tabular} &
\begin{tabular}[c]{@{}c@{}}\textbf{With}\\\textbf{VQA}\end{tabular} \\
\hline
nuPlan & 2021 & Real & Single-vehicle & 73 & \cmark & \xmark & \cmark & \cmark & \xmark \\
nuScenes & 2019 & Real & Single-vehicle & - & \xmark & \xmark & \xmark & \cmark & \xmark \\
CARLA LB V2 & 2023 & Sim. & Single-vehicle & 21 & \cmark & \cmark & \cmark & \cmark & \xmark \\
Bench2Drive & 2024 & Sim. & Single-vehicle & 44 & \cmark & \cmark & \cmark & \cmark & \xmark \\
WIBAM & 2021 & Real & Infrastructure & 1 & \xmark & \xmark & \cmark & \xmark & \xmark \\
V2X-Sim 2.0 & 2022 & Sim. & V2X & - & \xmark & \xmark & \xmark & \xmark & \xmark \\
OPV2V & 2021 & Sim. & V2X & 6 & \xmark & \xmark & \xmark & \xmark & \xmark \\
DAIR-V2X & 2021 & Real & V2X & 1 & \xmark & \xmark & \cmark & \cmark & \xmark \\
V2X-Seq & 2023 & Real & V2X & 1 & \xmark & \xmark & \cmark & \cmark & \xmark \\
V2Xverse & 2025 & Sim. & V2X & 24 & \cmark & \cmark & \cmark & \cmark & \xmark \\
\hline
\textbf{V2XBench} & 2026 & Sim. & V2X & 38 & \cmark & \cmark & \cmark & \cmark & \cmark \\
\hline
\end{tabular*}
\end{table*}

\section{Introduction}

Autonomous driving has witnessed remarkable progress in recent years; however, single-agent autonomous systems remain constrained by limited visibility and severe occlusions, particularly at complex intersections~\cite{prakash2021,shao2023}. To overcome these bottlenecks, Vehicle-to-Everything (V2X) cooperation has emerged as a promising paradigm~\cite{wang2020v2vnet,liu2025toward,xu2022opv2v,yu2022dair}. By facilitating the exchange of complementary observations among vehicles and roadside infrastructure, V2X-AD equips individual vehicles with an enriched beyond-line-of-sight perception of their surroundings. This collaborative approach expands the perceptive field, enabling intelligent agents to see beyond obstructions, anticipate hazards earlier, and mitigate safety risks in complex driving scenarios~\cite{xu2022v2xvit,hu2022where2comm}.

Despite this progress, current V2X research predominantly focuses on optimizing module-level capabilities, such as 3D object detection or semantic segmentation~\cite{li2022v2xsim,lu2024extensible,hu2023collaboration}. Most existing V2X systems remain constrained within traditional modular pipelines, where front-end perception and downstream planning are distinctly isolated~\cite{liu2025toward,cui2022coopernaut}. Consequently, these systems lack advanced semantic understanding and human-level logical reasoning. The absence of deep, high-level semantic interaction prevents these frameworks from seamlessly translating enriched collaborative perceptual data into comprehensive driving plans, ultimately limiting the evolution of V2X towards higher-order, fully autonomous driving.

Concurrently, Large Language Models (LLMs) and Vision-Language Models (VLMs) are reshaping autonomous driving with remarkable common-sense reasoning and trajectory planning capabilities~\cite{xu2024drivegpt4,sima2024drivelm,chen2025solve,ma2025autovla,cui2024survey,yang2024llm4drive,zhou2024vlm}. However, existing VLM-based models are restricted to an ``ego-only'' perspective, compromising their reasoning reliability under severe visual occlusions. We argue that extending VLMs to cooperative driving requires more than attaching a conventional fusion module: it calls for both benchmark infrastructure and feature alignment paradigm. Together, these components would enable VLMs to reason over globally informed V2X representations and translate cooperative observations into driving plans.

To address these requirements, we first introduce V2XBench, a CARLA-based V2X data infrastructure and simulation platform. V2XBench supports both offline cooperative model training and online closed-loop evaluation, utilizing strategically deployed roadside units (RSUs) to provide continuous coverage and eliminate ego-vehicle blind spots. Building on its synchronized multi-modal data and comprehensive annotations, we further construct Chat-V2XBench, a tailored Visual Question Answering (VQA) dataset for language-grounded cooperative driving. Its progressive QA hierarchy is explicitly designed to guide VLMs in mastering cross-view spatial alignment and cooperative end-to-end reasoning.

Building on this infrastructure, we propose AURORA (\textbf{A}utonomous driving via \textbf{U}nified \textbf{R}oadside-\textbf{O}riented \textbf{R}easoning \textbf{A}gent), an end-to-end framework for V2X cooperative driving. To reconcile the geometric and semantic discrepancies between ego and roadside views, AURORA introduces a query-level Cross-View Query Alignment and Fusion (CQAF) module. Rather than relying on rigid geometric pooling, CQAF transforms roadside queries into the ego coordinate frame, associates spatially corresponding queries, and fuses complementary semantic tokens into unified cooperative representations. A LoRA-adapted VLM then reasons over these representations, with the hidden state of a dedicated waypoint token conditioning a generative planner to produce future trajectories. In this way, AURORA connects cooperative perception, semantic reasoning, and closed-loop trajectory planning within a unified framework.

In summary, our contributions are threefold:
\begin{itemize}
\item We introduce V2XBench, a CARLA-based V2X platform supporting synchronized ego--roadside data generation and closed-loop evaluation, together with Chat-V2XBench, a progressively structured VQA dataset for language-grounded cooperative reasoning.
\item We propose AURORA, an end-to-end cooperative driving framework that employs CQAF to align and fuse ego and roadside queries and connects a LoRA-adapted VLM with a generative trajectory planner through a dedicated waypoint token.
\item We conduct extensive open- and closed-loop experiments to evaluate cooperative perception and driving performance. The results demonstrate the effectiveness of AURORA, particularly in occluded scenarios, while maintaining low roadside communication bandwidth.
\end{itemize}

\section{Related Work}

\subsection{V2X Simulation and Datasets}
The development of Vehicle-to-Everything (V2X) autonomous driving has been largely driven by the evolution of specialized datasets and simulation platforms. Real-world datasets such as DAIR-V2X~\cite{yu2022dair} and V2V4Real~\cite{xu2023v2v4real} provide synchronized multi-agent sensor observations for cooperative 3D perception. To reduce the cost and safety risks of real-world testing, simulation platforms such as OPV2V~\cite{xu2022opv2v} and V2X-Sim~\cite{li2022v2xsim} offer scalable environments, primarily for vehicle-to-vehicle perception. More recently, V2Xverse~\cite{liu2025toward} has extended V2X simulation to closed-loop end-to-end cooperative driving with diverse safety-critical scenarios. Despite this progress, existing V2X benchmarks largely focus on cooperative perception or closed-loop control, without language-grounded annotations and evaluation tasks for VLM-based cooperative reasoning. In contrast, V2XBench combines synchronized ego--roadside data generation with closed-loop evaluation, while Chat-V2XBench provides progressively structured VQA supervision for cross-view reasoning and planning.

\subsection{End-to-End Autonomous Driving and VLMs}
End-to-end autonomous driving maps raw sensor inputs directly to control signals or planning waypoints, mitigating error propagation of traditional modular designs. Representative single-agent methods, such as LAV~\cite{chen2022lav} and TCP~\cite{wu2022tcp}, have demonstrated strong closed-loop driving performance but remain limited by the visibility of onboard sensors. Concurrently, the integration of Large Language Models (LLMs) and Vision-Language Models (VLMs) into autonomous driving (e.g., DriveGPT4~\cite{xu2024drivegpt4}, DriveLM~\cite{sima2024drivelm}, LINGO-1~\cite{wayve2023lingo}) has introduced common-sense reasoning and scene-understanding capabilities. However, most existing driving VLMs operate from an ego-centric perspective, limiting their access to critical evidence when road users or hazards are occluded. To overcome the ego-centric bottleneck, our work pioneers a unified adaptation framework that extends VLM-based end-to-end reasoning into the V2X multi-agent domain.

\subsection{Cooperative Fusion Methods in V2X}
A central challenge in V2X-AD is effectively aggregating information across agents. To balance the limited information exchange of late fusion with the high communication cost of early fusion, intermediate (feature-level) fusion has become the mainstream paradigm. Representative methods such as F-Cooper~\cite{chen2019fcooper}, V2X-ViT~\cite{xu2022v2xvit}, and Where2comm~\cite{hu2022where2comm} aggregate features via spatial attention or point-pillar representations. More recent end-to-end methods, including Coopernaut~\cite{cui2022coopernaut}, UniV2X~\cite{yu2025univ2x}, and UniMM-V2X~\cite{unimmv2x}, further connect cooperative representations with driving-oriented perception and planning. However, their fused representations are primarily optimized for task-specific heads rather than explicitly structured as an interface for VLM-based reasoning. In contrast, CQAF performs geometry-aware alignment of cross-view spatial queries and subsequently injects the fused queries into semantic tokens, producing unified cooperative representations for VLM-guided trajectory planning.

\section{V2XBench: A Closed-Loop Platform for End-to-End Cooperative Autonomous Driving}

\begin{figure*}[t]
\centering
\includegraphics[width=\textwidth]{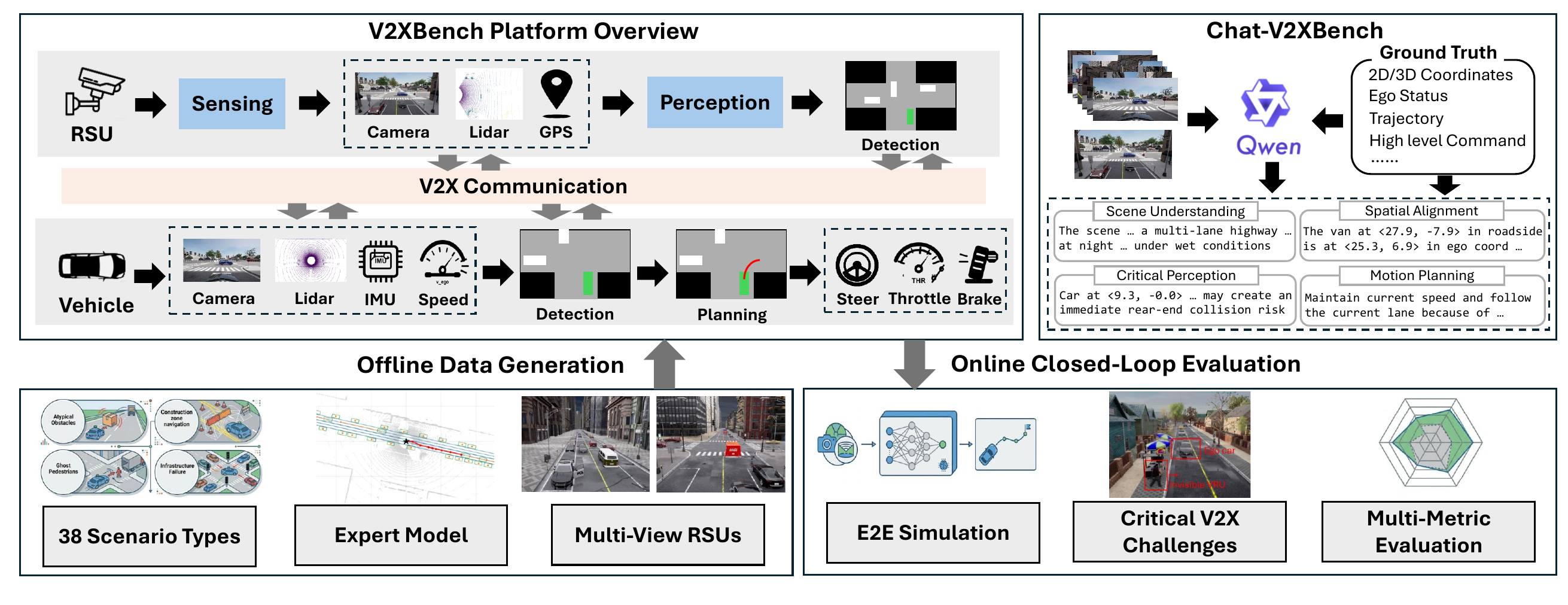}
\caption{Platform overview of V2XBench. V2XBench integrates route-adaptive roadside infrastructure, synchronized ego--RSU sensing, offline multi-agent data generation, Chat-V2XBench VQA construction, and online closed-loop cooperative driving evaluation within a unified CARLA-based simulation platform.}
\label{fig:v2xbench-platform}
\end{figure*}

To address the limitations of existing V2X benchmarks, we introduce V2XBench, a comprehensive simulation platform built atop the CARLA simulator~\cite{carla}. As illustrated in Fig.~\ref{fig:v2xbench-platform}, V2XBench provides a closed-loop ecosystem for end-to-end cooperative autonomous driving, consisting of four core components: 
(i) a  simulation environment featuring dual-link ego-RSU perception; 
(ii) Chat-V2XBench, a progressive VQA dataset designed to facilitate cooperative reasoning; 
(iii) an automated offline pipeline for synchronized multi-modal, multi-agent data generation; and 
(iv) a rigorous online protocol for closed-loop evaluation. 
In the following, we present the technical details of each component.

\subsection{Platform Construction}

V2XBench is constructed within the CARLA simulator, spanning multiple high-fidelity urban maps with diverse road topologies. The platform incorporates 38 safety-critical scenarios, strategically emphasizing complex multi-agent interactions and severe intersection occlusions where single-agent perception is fundamentally insufficient. To guarantee continuous infrastructure support, we introduce a route-adaptive RSU deployment strategy along every driving route. Each RSU is equipped with a synchronized, multi-modal sensor suite consisting of RGB, depth, and LiDAR sensors to maximize frontal observability. Correspondingly, the ego vehicle features a surround-view camera rig and a roof-mounted 360$^{\circ}$ LiDAR, ensuring full panoramic coverage while maintaining an efficient data volume for downstream training.

To simulate realistic visual occlusions essential for V2X research, V2XBench introduces a heavy-occluder mechanism that injects large background vehicles like trucks and vans to obstruct lines of sight. Furthermore, we deliberately introduce unpredictable jaywalking vulnerable road users (VRUs), including pedestrians and cyclists, into each scenario. This design faithfully reproduces high-risk ``ghost probe'' situations, capturing the core safety challenges that motivate cooperative V2X deployment in the real world. See Appendix~\ref{app:platform} for full scenario, sensor, and RSU deployment details.

\subsection{Offline Data Generation}

We employ the privileged expert agent PDM-Lite~\cite{sima2024drivelm} to collect a large-scale offline dataset. The resulting V2XBench dataset comprises 500 unique scenario runs across all 38 scenario types, yielding approximately 140K synchronized frames and over 2.5 million 3D bounding boxes. At each simulation step, the platform captures multi-modal sensory streams (including RGB, depth, LiDAR, and semantic maps) from both the ego vehicle and the dynamically selected nearest RSU, alongside full calibration matrices for precise cross-view transformations. To facilitate robust perception learning, the platform provides three distinct sets of 3D ground-truth (GT) annotations: ego-centric, roadside-centric, and cross-view fused GT. Each set contains comprehensive bounding box attributes---including class labels, velocities, and spatial poses---tailored for cooperative training tasks. A complete breakdown of the recorded data types is provided in Appendix~\ref{app:platform}.

\subsection{Online Closed-Loop Evaluation}

Unlike most existing V2X datasets that only support open-loop evaluation---where predicted trajectories are compared against ground-truth recordings without actual vehicle interaction---V2XBench enables closed-loop evaluation in which the driving agent must produce real-time control commands (throttle, steer, brake) and face the compounding consequences of its own decisions within a reactive simulation environment. This closed-loop paradigm is essential for faithfully assessing cooperative driving performance, as the benefits of V2X information often manifest precisely in dynamic interactions (e.g., timely braking upon receiving occluded-pedestrian alerts from RSUs) that open-loop metrics fundamentally cannot capture. During evaluation, the RSU infrastructure operates identically to the offline setting: the nearest RSU ahead of the ego vehicle is dynamically selected at each tick, and its sensor data is streamed to the agent alongside ego observations. We adopt the standard evaluation metrics including Route Completion (RC), Infraction Score (IS), and their product Driving Score (DS = RC $\times$ IS). Furthermore, we incorporate the Efficiency and Comfortness metrics proposed in Bench2Drive~\cite{bench2drive} to comprehensively assess the kinematic smoothness and practical driving quality of the generated trajectories. Beyond these standard metrics, we additionally consider V2X-specific communication bandwidth consumption to evaluate the practical deployability of cooperative approaches under realistic infrastructure constraints. More details can be found in Appendix~\ref{app:closed-loop}.

\subsection{Chat-V2XBench: A V2X-Aware VQA Dataset}

To empower VLMs with V2X-specific reasoning, we construct Chat-V2XBench, a large-scale VQA dataset designed to train models in cross-view spatial logic and complete logical reasoning chains. The dataset features a carefully designed four-level progressive QA hierarchy: 
(1)~\emph{Scene Understanding}---generating holistic descriptions of the environment, including weather, traffic density, and signal states; 
(2)~\emph{Critical Perception}---localizing key objects and assessing safety impacts by fusing multi-view ego and roadside observations; 
(3)~\emph{Spatial Alignment}---inferring an object's ego-centric coordinates given its position in the roadside camera frame, explicitly teaching the model perspective transformation; and 
(4)~\emph{Motion Planning}---deducing appropriate driving behaviors based on the cooperative context and justifying these actions physically. 
Overall, Chat-V2XBench comprises approximately 88K QA pairs spanning this reasoning hierarchy across diverse driving conditions. The complete question bank and our semi-automatic annotation pipeline are detailed in Appendix~\ref{app:chatv2x}.
\section{AURORA: Cooperative End-to-End Driving Framework}

\begin{figure*}[t]
\centering
\includegraphics[width=\textwidth]{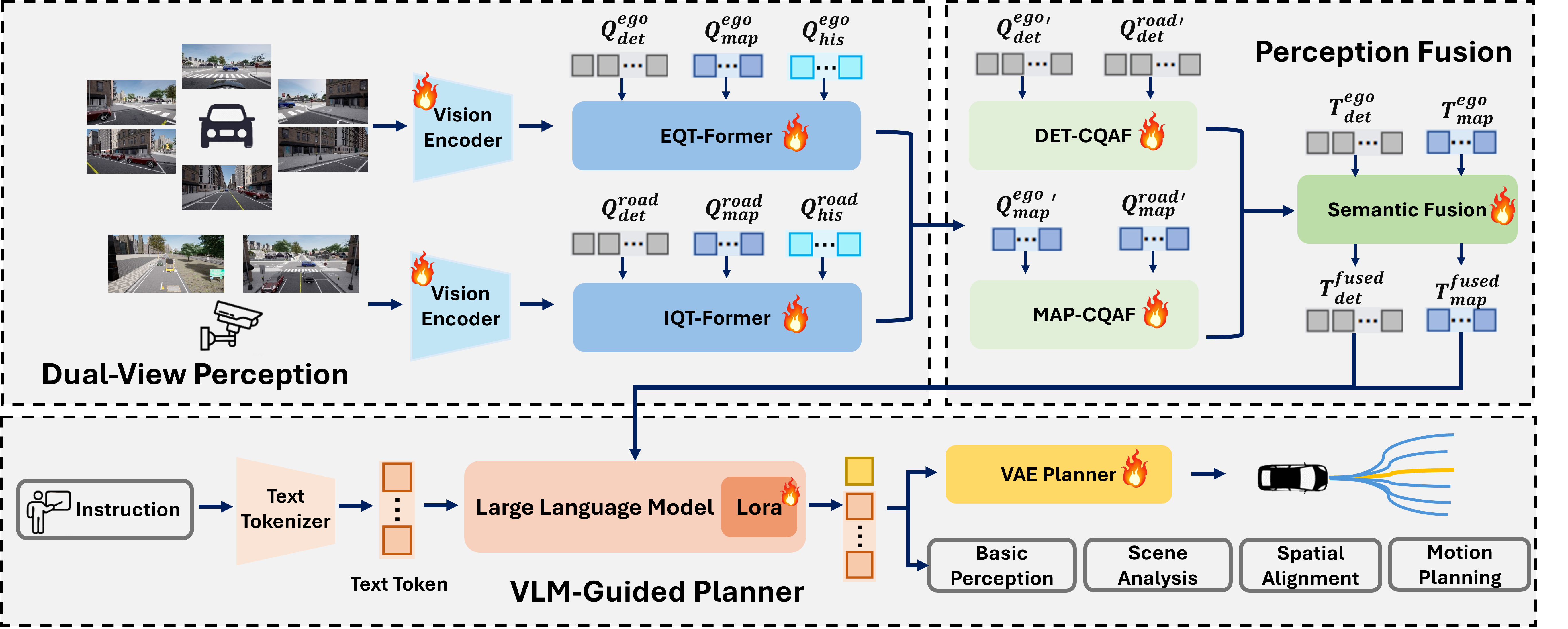}
\caption{The pipeline of our AURORA, a holistic end-to-end framework for V2X cooperative driving. The architecture first utilizes dual-view perception to encode synchronized ego and roadside streams into spatial queries and semantic tokens. Roadside detection and map queries are transmitted and fused in the ego frame by the Cross-View Query Alignment and Fusion (CQAF) module, while ego semantic tokens are refined via query-conditioned cross-attention. The resulting fused tokens ground a VLM-guided generative planner that bridges semantic reasoning and action to roll out future trajectories conditioned on a dedicated waypoint token.}
\label{fig:aurora-arch}
\end{figure*}

We propose AURORA (\textbf{A}utonomous driving via \textbf{U}nified \textbf{R}oadside-\textbf{O}riented \textbf{R}easoning \textbf{A}gent), a holistic end-to-end framework for V2X cooperative autonomous driving (Fig.~\ref{fig:aurora-arch}). 
The proposed architecture operates through a tightly integrated four-stage pipeline. 
First, dual-view backbones encode synchronized ego and roadside streams to extract high-quality detection and map queries (Sec.~4.1). 
Second, the Cross-View Query Alignment and Fusion (CQAF) module resolves perspective disparities by aligning and fusing spatial queries at the query level (Sec.~4.2). 
Third, the fused semantic tokens ground a LoRA-fine-tuned VLM, where the hidden state of a dedicated waypoint token conditions a generative planner to roll out future trajectories (Sec.~4.3). 
Finally, the entire framework is optimized end-to-end via a progressive multi-stage training curriculum (Sec.~4.4).

\subsection{Dual-View Perception}

Inspired by ViT~\cite{vit}, we construct two parameter-independent 3D encoders for the ego and roadside cameras, which RGB images as input and lift 2D image patches into 3D space using implicitly learned depth-aware positional embeddings tailored to their respective observational fields. Following ORION~\cite{orion}, each branch employs a detection decoder and a map decoder to process dynamic agents and static map elements. For a given view $v \in \{e,r\}$ (ego or roadside), these decoders emit spatial queries ($Q^{v}_{\mathrm{det}}$ and $Q^{v}_{\mathrm{map}}$) expressed in their own coordinate frames. Simultaneously, the ego branch distills its viewpoint features into semantic tokens ($T^{e}_{\mathrm{det}}$ and $T^{e}_{\mathrm{map}}$) that align directly with the language-model embedding space. To minimize transmission bandwidth, the roadside branch bypasses these dense semantic tokens and exclusively transmits its spatial queries. These complementary roadside queries are subsequently aligned and fused with the ego representations by the CQAF module.

\subsection{Cross-View Query Alignment and Fusion (CQAF)}

To systematically resolve spatial and semantic disparities between heterogeneous views, the CQAF module is structured into three dedicated sub-components: DET-CQAF for dynamic agents, MAP-CQAF for static map elements, and a unified semantic fusion for ego semantic tokens.

\noindent\textbf{DET-CQAF.}
We first align the roadside detection queries to the ego frame via the cross-view extrinsic transformation $E_{r\to e}$. Specifically, beyond transforming their 3D reference points, we explicitly concatenate the flattened rotation matrix of $E_{r\to e}$ to every roadside query before applying a cross-agent re-projection, ensuring strict positional and rotational compatibility. Based on spatial distances, we aggregate relevant roadside features via a temperature-controlled soft assignment and inject them into the ego stream using a learnable gated residual connection:
\begin{equation}
q^{f}_{\mathrm{det},i} = q^{e}_{\mathrm{det},i} + g_i \odot \mathrm{MLP}\bigl([\,q^{e}_{\mathrm{det},i};\, \tilde{q}^{r}_{\mathrm{agg},i}\,]\bigr),
\end{equation}
where $\tilde{q}^{r}_{\mathrm{agg},i}$ is the aggregated roadside feature and $g_i$ is a channel-wise gate. To recover objects outside the ego's field of view, we complete the fused set by appending distant, unmatched roadside queries. Rather than a simple geometric union, this dynamically enriches the ego representation with comprehensive roadside information supplementation, yielding the final fused query set $Q^{f}_{\mathrm{det}}$.

\noindent\textbf{MAP-CQAF.}
Since continuous lane polylines are ill-suited for point-wise soft matching, we instead filter roadside map queries by confidence. We then spatially align the valid survivors via $E_{r\to e}$ (encoded by a transformation function $\Phi$) and directly append them to the ego map set:
\begin{equation}
Q^{f}_{\mathrm{map}} = \bigl[\,Q^{e}_{\mathrm{map}};\; \Phi(E_{r\to e}\,Q^{r}_{\mathrm{map}})\bigr].
\end{equation}
This straightforward concatenation preserves topological continuity and supplies structural map features visible exclusively from the roadside infrastructure.

\noindent\textbf{Semantic fusion.}
The detection and map semantic token streams are processed independently. For each stream, the ego semantic tokens are refined via cross-attention keyed by their corresponding fused spatial queries ($Q^{f}_{\mathrm{det}}$ or $Q^{f}_{\mathrm{map}}$). This yields the fused semantic token sets $T^{f}_{\mathrm{det}}$ and $T^{f}_{\mathrm{map}}$, which serve as the consolidated inputs for the downstream VLM.

\subsection{VLM-Guided Generative Planner}

The concatenated fused tokens ($T^{f}_{\mathrm{det}}$ and $T^{f}_{\mathrm{map}}$) replace the visual placeholders of a LoRA-fine-tuned VLM, grounding its reasoning deeply in the cooperative perception. Guided by textual prompts, the model leverages Chat-V2XBench QA pairs to explicitly align the ego and roadside perspectives. To bridge semantic reasoning and continuous control, we introduce a dedicated waypoint token $\langle\mathrm{wp}\rangle$ into the tokenizer. Conditioned on the final-layer hidden state $h_{\mathrm{wp}}$ of this token, a probabilistic VAE planner samples a latent code from Gaussian distributions and rolls out candidate trajectories via a GRU decoder, with diffusion- and MLP-based alternatives detailed in Appendix~\ref{app:training}.

\subsection{Progressive Training Objective}

AURORA is trained via a five-stage curriculum. Stages~I and II independently pre-train the ego and roadside perception branches. Stage~III activates CQAF and the VLM, jointly optimizing cooperative perception and VQA alignment. Stage~IV unfreezes the planner under trajectory supervision, and Stage~V unfreezes all components for end-to-end refinement. The overall loss spans perception, cross-view fusion, language, and planning:
\begin{equation}
\begin{aligned}
\mathcal{L} \;=\;
& \underbrace{\Ldet{e} + \Lmap{e}}_{\text{ego perception}}
+ \underbrace{\Ldet{r} + \Lmap{r}}_{\text{roadside perception}} \\
& + \underbrace{\Ldet{f} + \Lmap{f}}_{\text{CQAF fused supervision}}
+ \Lvlm + \Lplan,
\end{aligned}
\label{eq:total-loss}
\end{equation}
where $\Ldet{*}$ and $\Lmap{*}$ denote standard set-prediction losses, and $\Lvlm$ is the next-token cross-entropy on the V2X-aware QA pairs. The planning term $\Lplan$ aggregates trajectory regression, lane-boundary, collision-avoidance, and KL-divergence penalties. We refer readers to Appendix~\ref{app:training} for the full stage-by-stage training details.

\section{Experiments}
\label{sec:experiments}

\begin{table*}[!t]
\centering
\caption{Cooperative 3D detection results on the V2XBench test split. All methods are trained and evaluated under the same ego--roadside cooperative setting. For clarity, we report class-wise mAP only for the main categories. BPS denotes bytes per second (B/s) for the roadside transmission cost.}
\label{tab:perception}
\small
\setlength{\tabcolsep}{4pt}
\begin{tabular*}{\textwidth}{@{\extracolsep{\fill}}lccGGccccc@{}}
\toprule
\multirow{2}{*}{\textbf{Method}} &
\multirow{2}{*}{\textbf{Reference}} &
\multirow{2}{*}{\textbf{Modality}} &
\multicolumn{1}{c}{\multirow{2}{*}{\textbf{Avg. mAP}\,\textuparrow}} &
\multicolumn{1}{c}{\multirow{2}{*}{\textbf{NDS}\,\textuparrow}} &
\multicolumn{4}{c}{\textbf{Main-class mAP}\,\textuparrow} &
\multirow{2}{*}{\begin{tabular}[c]{@{}c@{}}\textbf{Trans. Cost}\\\textbf{(BPS)}\,\textdownarrow\end{tabular}} \\
\cmidrule(lr){6-9}
& & & \multicolumn{1}{c}{} & \multicolumn{1}{c}{} &
\textbf{Vehicle} &
\textbf{VRUs} &
\textbf{Traffic Light} &
\textbf{Traffic Sign} &
\\
\midrule
No Fusion & -- & Camera & 0.478 & 0.497 & 0.513 & 0.532 & 0.204 & 0.301 & -- \\
\midrule
V2X-ViT & ECCV 22 & LiDAR & 0.350 & 0.485 & 0.521 & 0.042 & \textbf{0.530} & 0.397 & $2.10{\times}10^{8}$ \\
CoDriving & IEEE TPAMI 25 & LiDAR & 0.393 & 0.343 & 0.528 & \textbf{0.627} & 0.019 & 0.161 & $1.24{\times}10^{8}$ \\
UniV2X & AAAI 25 & Camera & 0.271 & 0.382 & 0.305 & 0.383 & 0.131 & 0.247 & $1.09{\times}\mathbf{10^{6}}$ \\
UniMM-V2X & AAAI 26 & Camera & 0.338 & 0.466 & 0.396 & 0.499 & 0.056 & 0.178 & $1.09{\times}\mathbf{10^{6}}$ \\
\midrule
\textbf{AURORA (Ours)} & -- & Camera & \textbf{0.548} & \textbf{0.568} & \textbf{0.582} & 0.605 & 0.290 & \textbf{0.400} & $4.84{\times}\mathbf{10^{6}}$ \\
\bottomrule
\end{tabular*}
\end{table*}

\begin{table*}[!t]
\caption{Planning results on the V2XBench test split under both closed-loop and open-loop evaluation. DS and RC denote Driving Score and Route Completion, respectively. BPS denotes bytes per second (B/s) for the roadside transmission cost.}
\label{tab:planning}
\centering
\small
\setlength{\tabcolsep}{4pt}
\begin{tabular*}{\textwidth}{@{\extracolsep{\fill}}lccGGccccc@{}}
\toprule
\multirow{2}{*}{\textbf{Method}} &
\multirow{2}{*}{\textbf{Reference}} &
\multirow{2}{*}{\textbf{Modality}} &
\multicolumn{4}{c}{\textbf{Closed-loop Metric}} &
\multicolumn{1}{c}{\textbf{Open-loop Metric}} &
\multirow{2}{*}{\begin{tabular}[c]{@{}c@{}}\textbf{Trans. Cost}\\\textbf{(BPS)}\,\textdownarrow\end{tabular}} \\
\cmidrule(lr){4-7}\cmidrule(lr){8-8}
& & &
\multicolumn{1}{c}{\textbf{DS}\,\textuparrow} &
\multicolumn{1}{c}{\textbf{RC}\,\textuparrow} &
\textbf{Efficiency}\,\textuparrow &
\textbf{Comfortness}\,\textuparrow &
\textbf{Avg. L2 (m)}\,\textdownarrow &
\\
\midrule
No Fusion & -- & Camera & 52.09 & 93.84 & 15.82 & 126.12 & 1.63 & -- \\
\midrule
CoDriving & IEEE TPAMI 25 & LiDAR & 56.75 & 91.23 & \textbf{18.59} & \textbf{193.55} & 1.38 & $1.24{\times}10^{8}$ \\
UniV2X & AAAI 25 & Camera & 33.76 & 86.76 & 13.1 & 128.82 & 2.19 & $4.05{\times}\mathbf{10^{6}}$ \\
UniMM-V2X & AAAI 26 & Camera & 49.26 & 87.33 & 5.2 & 74.55 & \textbf{1.03} & $4.66{\times}\mathbf{10^{6}}$ \\
\midrule
\textbf{AURORA (Ours)} & -- & Camera & \textbf{76.02} & \textbf{98.21} & 16.60 & 124.54 & 1.60 & $4.84{\times}\mathbf{10^{6}}$ \\
\bottomrule
\end{tabular*}
\end{table*}

\subsection{Baselines and Datasets}
\label{sec:baselines}
We benchmark AURORA against four representative V2X methods on our V2XBench dataset:
\begin{itemize}
\item \textbf{V2X-ViT}~\cite{xu2022v2xvit} fuses heterogeneous vehicle and infrastructure features at the intermediate level with a unified vision transformer.
\item \textbf{CoDriving}~\cite{liu2025toward} is an end-to-end collaborative driving system with a driving-oriented communication strategy; we adopt their official open-sourced LiDAR-input version.
\item \textbf{UniV2X}~\cite{yu2025univ2x} is a planning-oriented end-to-end cooperative framework using a sparse-dense hybrid transmission and fusion mechanism.
\item \textbf{UniMM-V2X}~\cite{unimmv2x} performs multi-level cooperation via shared queries and a Mixture-of-Experts architecture.
\end{itemize}

All baselines are re-trained from scratch on the V2XBench training split with their official configurations. The four V2X baselines are trained on $8\times$ NVIDIA A100 GPUs, while AURORA is trained on $8\times$ Kunlunxin~P800 GPUs. All models are then evaluated on the V2XBench benchmark.

\subsection{Perception Results}
\label{sec:perception}

To evaluate 3D perception, we adopt the standard nuScenes evaluation protocol, reporting mean Average Precision (mAP) via 2D center-distance matching and the nuScenes Detection Score (NDS). Specifically, NDS serves as a comprehensive metric calculated as a weighted combination of mAP and five True Positive errors (translation, scale, orientation, velocity, and attribute). While our dataset taxonomy encompasses a broader range of cooperative driving elements than standard nuScenes, we selectively report mAP for four primary categories, with VRUs obtained by averaging pedestrian and bicycle mAP.

As shown in Table~\ref{tab:perception}, AURORA establishes a new state-of-the-art among V2X cooperative methods. It significantly outperforms existing collaborative baselines, achieving absolute gains of +0.155 in Avg.\ mAP (over CoDriving) and +0.083 in NDS (over V2X-ViT). Remarkably, as a camera-based architecture, AURORA surpasses heavily parameterized LiDAR-based methods in overall NDS, validating the effectiveness of our CQAF module in achieving precise cross-view alignment at the high-dimensional query level without dense point cloud transmission. In class-wise performance, AURORA attains the highest mAP for Vehicles (0.582) and Traffic Signs (0.400), while maintaining competitive accuracy across vulnerable road users and traffic lights, thereby securing a robust semantic foundation for downstream planning.

\subsection{Planning Results}
\label{sec:planning}

To comprehensively evaluate the end-to-end driving capabilities, we assess the models using both closed-loop interactive metrics (Driving Score, Route Completion, Efficiency, and Comfortness) and the standard open-loop trajectory error (Avg.\ L2 distance). For closed-loop execution, CoDriving utilizes its native controller. Since UniV2X and UniMM-V2X were originally proposed on open-loop datasets, we equip them with the same controller as AURORA to ensure a fair comparison.

As summarized in Table~\ref{tab:planning}, AURORA establishes a new state-of-the-art in closed-loop cooperative driving across all evaluated baselines. It achieves an outstanding Driving Score (DS) of 76.02 and a Route Completion (RC) rate of 98.21\%, significantly outperforming the best-performing baseline, CoDriving, by a large margin (+19.27 in DS and +6.98\% in RC). Interestingly, while UniMM-V2X achieves the lowest open-loop L2 error (1.03\,m), its closed-loop DS (49.26) and Efficiency (5.2) degrade severely in the simulation environment. This underscores a critical paradigm gap: pure trajectory regression optimized for open-loop settings often fails to translate into reactive and safe driving in dynamically complex scenarios. By grounding generative planning in VLM-based semantic reasoning, AURORA successfully bridges this gap, yielding highly robust interactive behaviors. Crucially, AURORA maintains the bandwidth advantage of query-level fusion, requiring nearly two orders of magnitude less transmission cost than methods like CoDriving.

\subsection{Ablation Study}
\label{sec:ablation}

We conduct extensive ablation studies on the V2XBench dataset to validate the core components of AURORA.

\paragraph{Effectiveness of Cross-View Fusion.}
As shown in Table~\ref{tab:ablation-fusion}, we first ablate the ego--roadside fusion mechanism by comparing AURORA against an ego-only baseline. Incorporating the CQAF module yields substantial improvements in cooperative 3D perception, increasing mAP by $0.070$ and NDS by $0.071$. More importantly, this enhanced environmental awareness translates directly into superior downstream planning, boosting the closed-loop Driving Score (DS) from 52.09 to 76.02. This confirms the critical role of roadside cooperative queries in mitigating visual occlusions and ensuring safe navigation.

\begin{table}[!ht]
\centering
\caption{Ablation on ego--roadside fusion. DS and RC denote Driving Score and Route Completion. ``w/'' and ``w/o'' denote with and without.}
\label{tab:ablation-fusion}
\small
\setlength{\tabcolsep}{2.5pt}
\begin{tabular}{@{}lccccc@{}}
\toprule
\multirow{2}{*}{\textbf{Setting}} &
\multicolumn{2}{c}{\textbf{Perception}} &
\multicolumn{2}{c}{\textbf{Closed-loop}} &
\multicolumn{1}{c}{\textbf{Open-loop}} \\
\cmidrule(lr){2-3}\cmidrule(lr){4-5}\cmidrule(lr){6-6}
& \textbf{mAP}\,\textuparrow & \textbf{NDS}\,\textuparrow &
\textbf{DS}\,\textuparrow & \textbf{RC}\,\textuparrow &
\textbf{Avg. L2 (m)}\,\textdownarrow \\
\midrule
w/o Fusion & 0.478 & 0.497 & 52.09 & 93.84 & 1.63 \\
\midrule
\textbf{w/ Fusion (Ours)} & 0.548 & 0.568 & 76.02 & 98.21 & 1.60 \\
\bottomrule
\end{tabular}
\end{table}
\FloatBarrier

\paragraph{Planner Architecture.}
Table~\ref{tab:ablation-planner} evaluates the impact of different generative planning decoders. While the Diffusion planner clearly outperforms the deterministic MLP decoder, our probabilistic VAE planner achieves the best overall performance with a DS of 76.02 and a Route Completion (RC) of 98.21\%, alongside the lowest collision rate (3.47\%). By effectively modeling the multimodal distribution of future states via latent Gaussian sampling, the VAE planner demonstrates superior robustness in complex, reactive closed-loop environments.

\begin{table}[!ht]
\centering
\caption{Ablation on generative planners. DS and RC denote Driving Score and Route Completion.}
\label{tab:ablation-planner}
\small
\setlength{\tabcolsep}{3.5pt}
\begin{tabular}{@{}lcccc@{}}
\toprule
\multirow{2}{*}{\textbf{Planner}} &
\multicolumn{2}{c}{\textbf{Closed-loop}} &
\multicolumn{2}{c}{\textbf{Open-loop}} \\
\cmidrule(lr){2-3}\cmidrule(lr){4-5}
& \textbf{DS}\,\textuparrow & \textbf{RC}\,\textuparrow & \textbf{Avg. L2 (m)}\,\textdownarrow & \textbf{Avg. Col. (\%)}\,\textdownarrow \\
\midrule
MLP & 56.98 & 88.38 & 1.94 & 5.18 \\
Diffusion & 66.21 & 92.61 & 1.73 & 3.91 \\
VAE (Ours) & 76.02 & 98.21 & 1.60 & 3.47 \\
\bottomrule
\end{tabular}
\end{table}
\FloatBarrier

\paragraph{Impact of VQA Supervision.}
Table~\ref{tab:ablation-vqa} investigates the necessity of Chat-V2XBench VQA supervision during VLM adaptation. Although removing the VQA loss slightly reduces the open-loop L2 error (1.54\,m vs.\ 1.60\,m)---a common phenomenon when models overfit to pure trajectory regression without auxiliary semantic constraints---it severely degrades interactive driving capabilities. Without language-grounded reasoning, the closed-loop DS drops by over 8 points (from 76.02 to 67.98). This demonstrates that the progressive VQA supervision is indispensable for empowering the VLM with high-level spatial and logical reasoning, which fundamentally dictates survival in safety-critical scenarios.

\begin{table}[!ht]
\centering
\caption{Ablation on VQA supervision. DS and RC denote Driving Score and Route Completion. ``w/'' and ``w/o'' denote with and without.}
\label{tab:ablation-vqa}
\small
\setlength{\tabcolsep}{3.5pt}
\begin{tabular}{@{}lcccc@{}}
\toprule
\multirow{2}{*}{\textbf{Setting}} &
\multicolumn{2}{c}{\textbf{Closed-loop}} &
\multicolumn{2}{c}{\textbf{Open-loop}} \\
\cmidrule(lr){2-3}\cmidrule(lr){4-5}
& \textbf{DS}\,\textuparrow & \textbf{RC}\,\textuparrow & \textbf{Avg. L2 (m)}\,\textdownarrow & \textbf{Avg. Col. (\%)}\,\textdownarrow \\
\midrule
w/o VQA & 67.98 & 92.11 & 1.54 & 3.57 \\
w/ VQA (Ours) & 76.02 & 98.21 & 1.60 & 3.47 \\
\bottomrule
\end{tabular}
\end{table}

\section{Conclusion}
In this paper, we address the limited support for closed-loop evaluation and language-grounded supervision in cooperative autonomous driving by introducing the V2XBench simulation platform and the Chat-V2XBench VQA dataset. Building upon this infrastructure, we propose AURORA, a novel end-to-end cooperative driving framework. AURORA bridges ego and roadside perspectives through the Cross-View Query Alignment and Fusion (CQAF) module and leverages the resulting cooperative representations to connect VLM-based semantic reasoning with safe, executable trajectory generation. Extensive evaluations demonstrate that AURORA achieves state-of-the-art performance in both cooperative 3D perception and closed-loop driving. Despite relying solely on camera inputs, it surpasses LiDAR-based baselines in overall detection accuracy. More importantly, AURORA effectively translates enhanced cooperative perception into safe navigation under occlusion-heavy scenarios, narrowing the gap between open-loop trajectory prediction and reactive closed-loop driving. These improvements are achieved while maintaining low query-level communication bandwidth. Ultimately, this work pioneers a scalable, VLM-driven paradigm for next-generation cooperative autonomous systems.

\bibliography{aaai2027}

\clearpage
\appendix
\section{V2XBench Dataset Details}
\label{app:v2xbench}
\subsection{Platform Construction}
\label{app:platform}

V2XBench is built on top of the CARLA simulator and the CARLA Leaderboard~2.0 scenario suite. Data are collected offline by rolling out a privileged expert agent (PDM-Lite~\cite{sima2024drivelm}) in closed-loop mode, while a route-adaptive roadside sensing system records synchronized ego and infrastructure observations. The scenarios are instantiated across eight CARLA towns (Town01--Town05, Town10HD, and the large-scale Town12/Town13), with the two large maps contributing the majority of runs to maximize geometric and semantic diversity. In total, the platform yields $500$ scenario runs, producing roughly $140$K synchronized frames.

\subsubsection{Sensor Configuration.}
Table~\ref{tab:sensors} summarizes the sensor suite. The ego vehicle carries a $360^\circ$ surround rig of six pinhole cameras mounted at a height of $1.8$\,m, oriented at yaw angles of $0^\circ$ (front), $\pm60^\circ$ (front-left/right), $\pm120^\circ$ (rear-left/right), and $180^\circ$ (rear). Each viewpoint simultaneously provides spatially aligned RGB, depth, and semantic-segmentation images at $1024\times512$ resolution with a $110^\circ$ field of view (FOV). A roof-mounted spinning LiDAR ($z{=}2.5$\,m) operates at $10$\,Hz, and a rasterized bird's-eye view (BEV) semantic map is rendered as auxiliary ground truth.

Each roadside unit (RSU) comprises a co-located RGB and depth camera pair ($1024\times512$, $90^\circ$ FOV) and a $64$-beam LiDAR (up to $1.0$M points/s, $75$\,m range, $120^\circ$ horizontal FOV and a $[-45^\circ, 15^\circ]$ vertical FOV). Every RSU is elevated to $5$\,m and pitched slightly downward toward the roadway (with a small roll offset) so as to overlook the conflict zone, emulating a realistic infrastructure-mounted perspective.

\begin{table}[!ht]
\centering
\caption{Sensor suite of the ego vehicle and each roadside unit (RSU) in V2XBench. All cameras additionally provide spatially aligned depth and (for the ego) semantic-segmentation streams.}
\label{tab:sensors}
\small
\setlength{\tabcolsep}{4pt}
\begin{tabular}{@{}lll@{}}
\toprule
\textbf{Sensor} & \textbf{Configuration} & \textbf{Mounting} \\
\midrule
\multicolumn{3}{@{}l}{\textit{Ego vehicle}}\\
$6\times$ RGB cam.    & $1024{\times}512$, FOV $110^\circ$ & surround, $z{=}1.8$\,m \\
$6\times$ Depth cam.  & $1024{\times}512$, FOV $110^\circ$ & paired with RGB \\
$6\times$ Sem.\ cam.  & $1024{\times}512$, FOV $110^\circ$ & paired with RGB \\
$1\times$ LiDAR       & $10$\,Hz, $0.6$M pts/s             & roof, $z{=}2.5$\,m \\
BEV semantics         & rasterized GT map                  & --- \\
\midrule
\multicolumn{3}{@{}l}{\textit{Roadside unit (RSU)}}\\
$1\times$ RGB cam.    & $1024{\times}512$, FOV $90^\circ$  & $z{=}5$\,m, downward \\
$1\times$ Depth cam.  & $1024{\times}512$, FOV $90^\circ$  & paired with RGB \\
$1\times$ LiDAR       & $64$\,ch, $1.0$M pts/s, $75$\,m    & H-FOV $120^\circ$ \\
\bottomrule
\end{tabular}
\end{table}

A defining feature of V2XBench is its \emph{route-adaptive roadside deployment}, which guarantees that at every timestep the ego vehicle is served by an active RSU providing effective beyond-line-of-sight information. Instead of fixing infrastructure at a few hand-picked poses, we densely distribute candidate RSUs along each route: at regular intervals ($\sim$$80$\,m) along straight segments, at every intersection (offset from the junction center to overlook all incoming approaches), and near the trigger point of each safety-critical event so that the conflict is observed before it unfolds, while pruning redundant units that are too close together. At run time, for each synchronized frame the system selects the nearest forward-facing RSU as the active infrastructure view, ensuring continuous and informative cross-view coverage throughout the route.

\subsubsection{Scenario Types.}
\begin{table*}[t]
\centering
\caption{The $38$ safety-critical scenario types in V2XBench, grouped into V2X scenarios (occlusion- and beyond-line-of-sight-dominated, where roadside infrastructure is most beneficial) and other challenging autonomous-driving scenarios.}
\label{tab:scenario-types}
\small
\begin{tabular}{@{}>{\raggedright\arraybackslash}p{0.16\textwidth}>{\raggedright\arraybackslash}p{0.78\textwidth}@{}}
\toprule
\textbf{Category} & \textbf{Scenario Types} \\
\midrule
V2X scenarios & CrossingBicycleFlow, DynamicObjectCrossing, NonSignalizedJunctionLeftTurn, NonSignalizedJunctionRightTurn, ParkingCrossingPedestrian, PedestrianCrossing, VehicleTurningRoute, VehicleTurningRoutePedestrian \\
\midrule
Other scenarios & Accident, AccidentTwoWays, BlockedIntersection, ConstructionObstacle, ConstructionObstacleTwoWays, ControlLoss, EnterActorFlow, EnterActorFlowV2, HardBreakRoute, HazardAtSideLane, HazardAtSideLaneTwoWays, HighwayCutIn, HighwayExit, InterurbanActorFlow, InterurbanAdvancedActorFlow, InvadingTurn, MergerIntoSlowTraffic, MergerIntoSlowTrafficV2, OppositeVehicleRunningRedLight, OppositeVehicleTakingPriority, ParkedObstacle, ParkedObstacleTwoWays, ParkingCutIn, ParkingExit, PriorityAtJunction, SignalizedJunctionLeftTurn, SignalizedJunctionRightTurn, StaticCutIn, VehicleOpensDoorTwoWays, YieldToEmergencyVehicle \\
\bottomrule
\end{tabular}
\end{table*}
\begin{figure*}[t]
\centering
\includegraphics[width=0.75\textwidth]{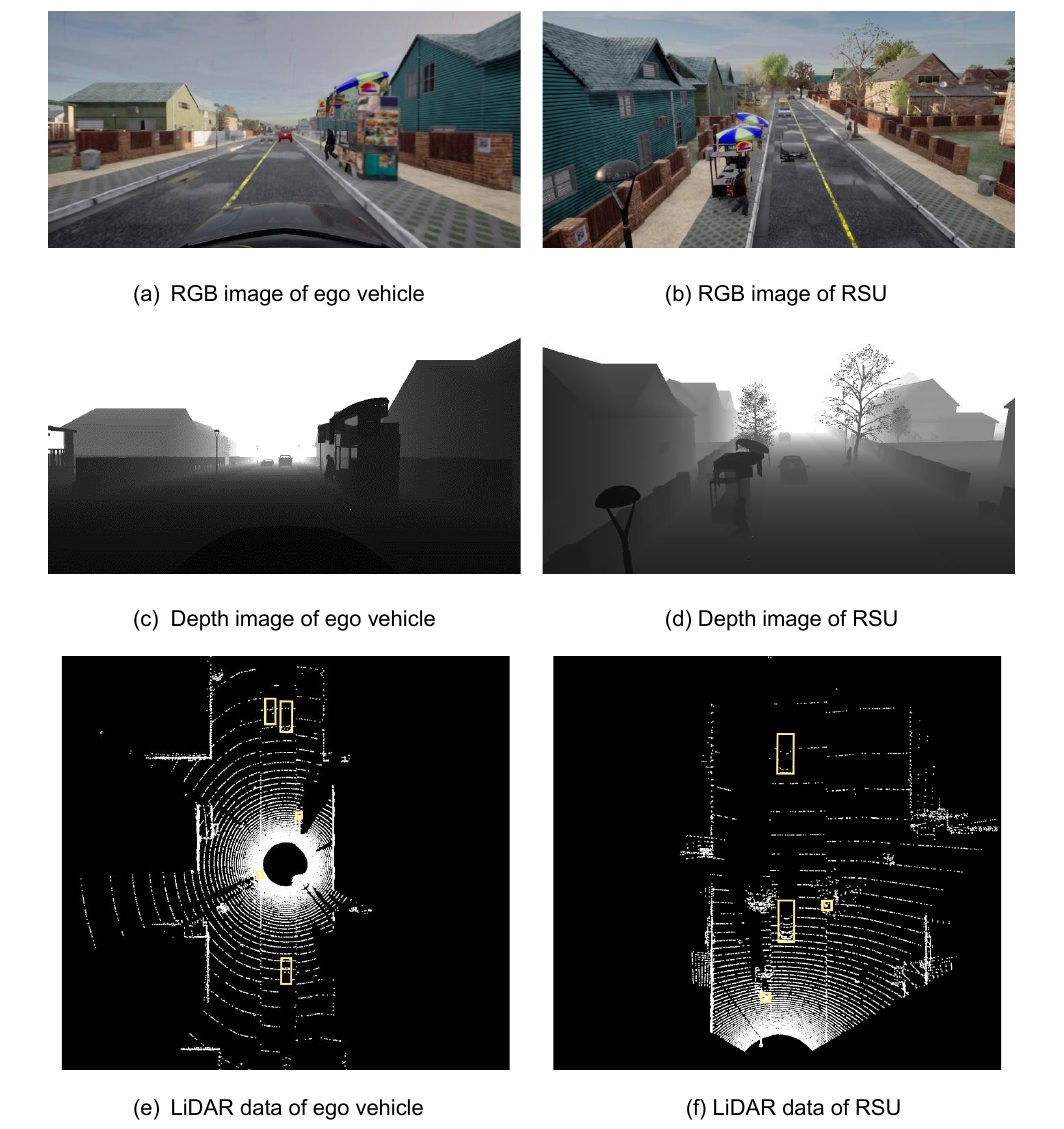}
\caption{Representative visualizations of synchronized V2XBench sensory data, including RGB images, spatially aligned depth maps, and LiDAR point clouds captured from ego and roadside views.}
\label{fig:data-types}
\end{figure*}
V2XBench covers $38$ safety-critical scenario types, grouped into two categories in Table~\ref{tab:scenario-types}. To directly tackle the core pain points of cooperative driving, we place particular emphasis on collecting \emph{V2X scenarios}, namely settings dominated by occlusion and beyond-line-of-sight hazards where a single ego agent fails and roadside infrastructure is decisive. Meanwhile, we retain a broad range of other challenging autonomous-driving scenarios to enrich the diversity of our data. To further amplify the density of V2X-relevant situations, in every route we additionally spawn multiple vulnerable road users (VRUs) crossing the roadway and raise the proportion of large vehicles (e.g., trucks and vans), both of which aggravate occlusions and thus artificially create more opportunities for cooperative perception.

\subsubsection{Data Types.}
At each simulation step, V2XBench records a rich set of synchronized, multi-modal data from both the ego vehicle and the active RSU, together with the full set of calibration and intrinsic/extrinsic matrices needed for precise cross-view transformation. In total, the platform yields roughly $140$K synchronized frames and over $2.5$ million annotated 3D bounding boxes. Figure~\ref{fig:data-types} visualizes representative examples of the core sensory modalities. The recorded data are organized into the following types:
\begin{itemize}
\item \textbf{Camera data.} Paired ego--roadside RGB images (six surround-view ego cameras and one active roadside camera). Captured from geometrically complementary vantage points, they constitute the primary sensory input and are the key enabler of the cross-view cooperative perception studied in this work.
\item \textbf{LiDAR data.} Synchronized ego and roadside point clouds that provide dense 3D geometry complementing the cameras and supporting LiDAR-based and multi-modal baselines.
\item \textbf{Depth data.} A spatially aligned, pixel-wise depth image accompanying every RGB camera, facilitating lifting 2D image features into 3D space and improving the geometric accuracy of cross-view perception.
\item \textbf{Semantic data.} Pixel-wise semantic-segmentation maps from the ego cameras together with a rasterized BEV semantic map, describing static scene layout (e.g., drivable area, lanes, sidewalks) and dynamic categories to supervise online mapping and scene understanding.
\item \textbf{Expert trajectories.} For every run we log the privileged expert (PDM-Lite~\cite{sima2024drivelm}) rollout, including the executed ego trajectory (poses and velocities) and the corresponding low-level control commands (throttle, steer, brake) at each tick. These expert trajectories serve as the imitation-learning targets for the planning task and as the reference behavior for closed-loop comparison.
\item \textbf{Bounding boxes.} The platform provides three distinct sets of 3D bounding-box annotations---ego-centric, roadside-centric, and cross-view fused---so that cooperative and single-agent settings can be trained and evaluated consistently. Each box carries complete attributes including class label, 3D pose (location, orientation, dimensions), and velocity. Beyond vehicles, large vehicles, and vulnerable road users (VRUs), the annotations also cover traffic-control elements such as traffic lights, traffic signs, and traffic cones, as well as other road objects that may appear in the scene.
\item \textbf{QA pairs.} Finally, each scene is annotated with V2X-aware question--answer pairs from Chat-V2XBench, our VQA dataset spanning a four-level progressive hierarchy (scene understanding, critical perception, spatial alignment, and motion planning). These QA pairs ground language reasoning in the cooperative visual context and are used to train and evaluate the VLM component; see Appendix~A.2 for details.
\end{itemize}

\subsection{Details on Chat-V2XBench}
\label{app:chatv2x}

\begin{figure*}[t]
\centering
\includegraphics[width=\textwidth]{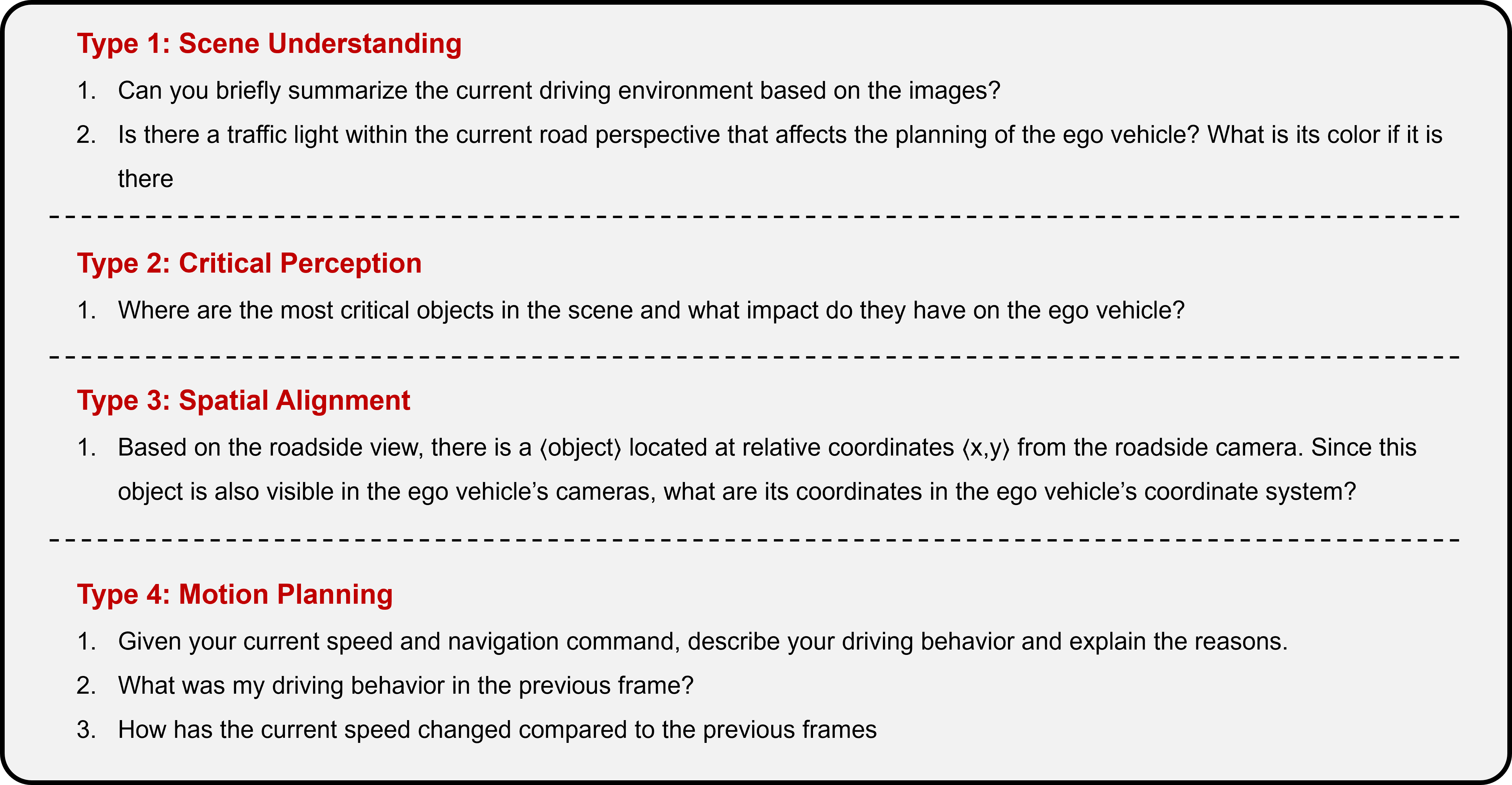}
\caption{The Chat-V2XBench question bank. Each keyframe is annotated with questions drawn from the four levels of the progressive reasoning hierarchy.}
\label{fig:qa-bank}
\end{figure*}

\begin{figure*}[t]
\centering
\includegraphics[width=\textwidth]{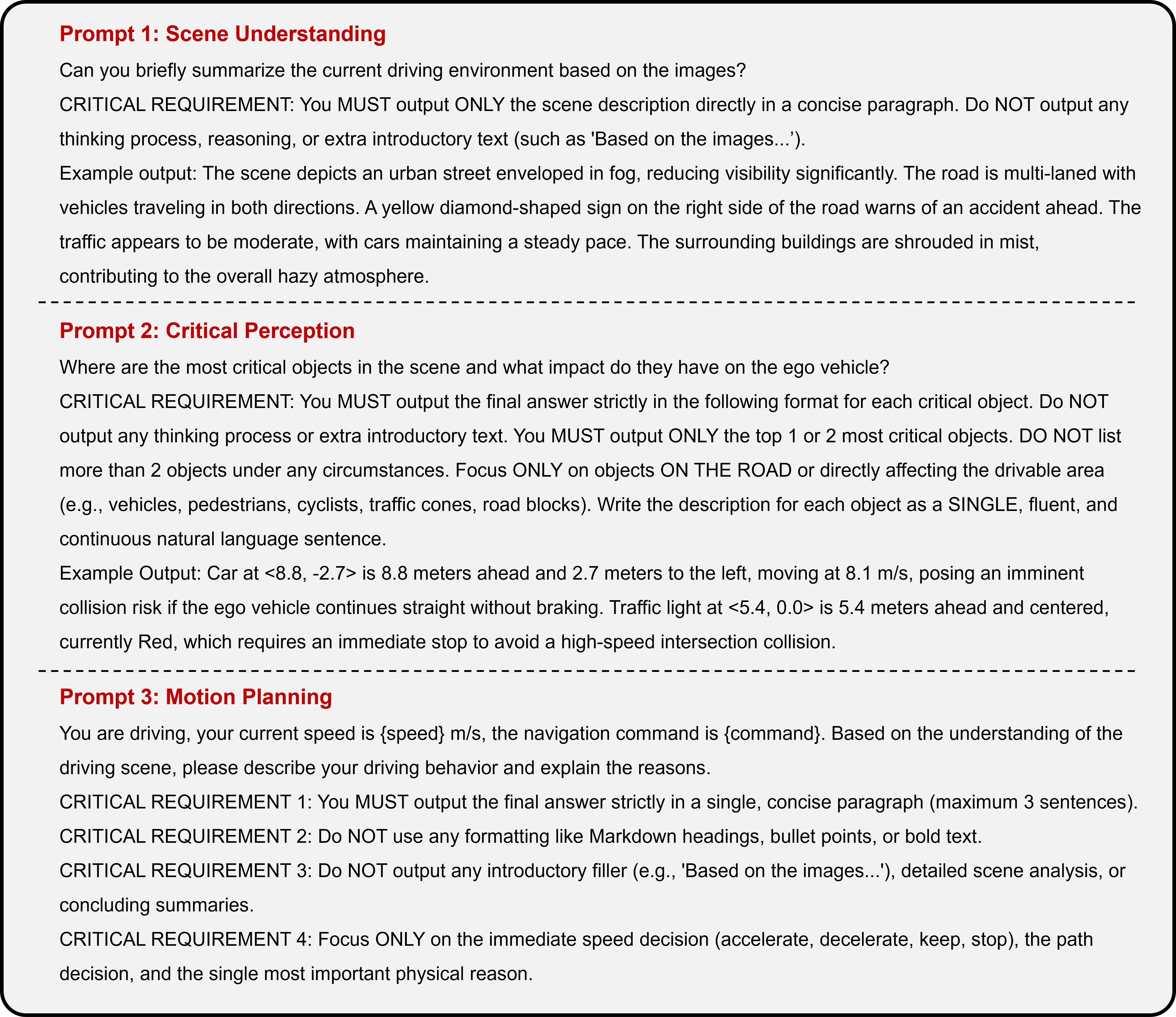}
\caption{The prompt used to auto-label the open-ended QA pairs with Qwen3-VL-Plus, conditioning the model on the multi-view imagery and the ground-truth object hints.}
\label{fig:anno-prompt}
\end{figure*}

Chat-V2XBench is the language-grounded layer of V2XBench, pairing every sampled keyframe with a set of V2X-aware question--answer (QA) pairs. Rather than annotating every simulation step, we sample one keyframe every $10$ frames---that is, one keyframe per second of simulation time given the $10$\,Hz recording rate---for QA annotation, which avoids near-duplicate labels on temporally adjacent frames while retaining sufficient temporal coverage. All QA instances are anchored on the same synchronized visual context---the six surround-view ego cameras together with the co-located roadside camera---so that answering inherently requires reasoning jointly over the ego and infrastructure viewpoints. To ground the language supervision in accurate geometry, every prompt is additionally conditioned on (i)~a compact textual list of nearby ground-truth objects with their ego-frame coordinates, class labels, and speeds, and (ii)~the relative position and heading of the roadside camera with respect to the ego vehicle. Below we first detail the QA content following the four-level progressive hierarchy, and then describe the semi-automatic annotation pipeline used to produce the answers.

\subsubsection{Question--Answer Content.}
Following the four-level hierarchy, each keyframe is annotated with questions spanning progressively harder cooperative-reasoning skills (the full question bank is shown in Fig.~\ref{fig:qa-bank}).

\begin{itemize}
\item \textbf{Scene Understanding.} At the coarsest level, the model is asked to produce a holistic description of the driving environment, covering weather and time of day, traffic density, road conditions (e.g., construction signs or obstacles on the road surface), and the prevailing traffic-signal states. A complementary closed-form query explicitly asks whether a traffic light currently affects the ego vehicle and, if so, what its color is. These questions establish a global context for all subsequent reasoning.
\item \textbf{Critical Perception.} The model must localize the one or two \emph{most critical} on-road objects by fusing the multi-view ego and roadside observations, report their ego-centric coordinates and motion, and articulate the concrete safety impact on the ego vehicle (e.g., imminent collision risk or the need to brake). 
\item \textbf{Spatial Alignment.} Given an object's coordinates expressed in the roadside camera frame, the model is asked to infer the same object's coordinates in the ego vehicle frame, explicitly teaching the cross-view perspective transformation that is central to cooperative driving. To guarantee that the target is genuinely observable from both sides, we only sample objects that are jointly visible, requiring a sufficient number of LiDAR returns ($\ge 10$ points) in both the ego and roadside sensors, and compute the answer from the exact ego-to-roadside rigid transformation.
\item \textbf{Motion Planning.} At the highest level, conditioned on the current speed and navigation command, the model must decide the immediate driving behavior---the speed decision (accelerate, decelerate, keep, or stop) together with the path decision---and justify it with the single most important physical reason grounded in the cooperative context. Two auxiliary short-horizon queries about the ego's own recent behavior (the previous-frame navigation command) and speed trend (accelerating, decelerating, or unchanged) further anchor the model's temporal self-awareness.
\end{itemize}

\subsubsection{Annotation Pipeline.}
Depending on whether a question admits a deterministic ground-truth answer, we adopt a two-track annotation strategy (the shared annotation prompt is shown in Fig.~\ref{fig:anno-prompt}).

\begin{itemize}
\item \textbf{Ground-truth answers.} Objective questions are answered directly from the simulator's privileged state, yielding exact labels free of annotation noise. This includes the traffic-light query (from the logged signal state and its affect-ego flag), the previous-frame behavior and speed-trend queries (from the logged navigation command and frame-to-frame speed difference), and the spatial-alignment query (from the exact rigid transformation between the ego and roadside coordinate frames). Because these answers are computed geometrically or read off the ground truth, they require no further verification.
\item \textbf{Auto-labeled answers with human audit.} Open-ended, descriptive questions---holistic scene understanding, critical-object reasoning, and behavior-with-justification---cannot be reduced to a template. We therefore first auto-label them with a strong vision--language model, Qwen3-VL-Plus, prompting it with the same multi-view images together with the ground-truth object hints so that its free-form responses remain consistent with the true scene geometry. We then conduct human spot-checking: annotators randomly sample the generated answers and cross-check them against the ground-truth state and imagery, correcting or discarding any response that is factually inconsistent, hallucinated, or physically unreasonable. Crucially, this auditing is iterative---based on the failure patterns surfaced during spot-checking, we continually refine the annotation prompt (e.g., tightening the output format, emphasizing the ground-truth object hints, and suppressing common hallucinations) and re-label the affected data, until the sampled VQA annotations reliably conform to the actual scene. This ground-truth-conditioned auto-labeling followed by iterative human verification provides fluent natural-language supervision while preserving factual reliability.
\end{itemize}

\subsection{Closed-Loop Evaluation Details}
\label{app:closed-loop}

To rigorously assess the end-to-end driving capabilities of the proposed framework, we construct a comprehensive closed-loop evaluation benchmark within the CARLA simulator. All closed-loop testing is conducted on Town12, a massive and highly realistic map chosen for its extensive scale and diverse road topologies, which naturally provide all types of complex driving environments ranging from dense urban intersections to multi-lane highways. Within this environment, we configure 50 distinct and challenging test routes that comprehensively encompass all corner-case scenarios defined in our dataset. Importantly, to prevent map overfitting and evaluate genuine cross-scenario generalization, these testing routes are strictly disjoint from the training split. To specifically stress-test the cooperative perception and reasoning limits of the models, we intentionally over-sample and strengthen V2X-dominated scenarios along these routes---such as heavily occluded non-signalized junctions and unpredictable pedestrian crossings---where single-vehicle perception is fundamentally constrained.

To provide a multifaceted assessment of the driving quality, we employ five quantitative metrics spanning safety, progress, kinematics, and system efficiency:

\begin{itemize}
    \item \textbf{Route Completion (RC):} The percentage of the predefined route distance successfully navigated by the ego vehicle before timing out or triggering a terminal failure.
    \item \textbf{Driving Score (DS):} The primary CARLA Leaderboard metric, calculated as the product of Route Completion (RC) and an Infraction Score (IS). The IS acts as a penalty multiplier (bounded between 0 and 1) that decays exponentially each time the ego vehicle commits a safety infraction, such as collisions with other agents or static layouts, red-light violations, or lane invasions.
    \item \textbf{Efficiency:} Following the Bench2Drive~\cite{bench2drive} protocol, Efficiency evaluates the kinematic progression of the agent. It measures the vehicle's ability to maintain an optimal and reasonable driving speed, penalizing overly conservative behaviors such as unnecessary stops or sluggish cruising when the forward road is clear.
    \item \textbf{Comfortness:} Also adapted from Bench2Drive~\cite{bench2drive}, Comfortness quantifies the kinematic smoothness of the generated trajectory. It is computed by tracking the vehicle's dynamic state and penalizing extreme longitudinal or lateral accelerations (e.g., sudden braking, hard acceleration, or sharp steering) as well as excessive jerk and yaw rates, thereby reflecting the ride quality for passengers.
    \item \textbf{Communication Bandwidth:} To evaluate the practical deployability of cooperative systems under real-world infrastructure constraints, we explicitly measure the transmission cost between the active roadside unit (RSU) and the ego vehicle. This metric is reported in BPS (bytes per second, B/s), quantifying the continuous data payload required to sustain the V2X cooperation during driving.
\end{itemize}

\section{AURORA Training Details}
\label{app:training}

\label{appendix:training_details}

We train AURORA with a five-stage curriculum. Each stage unlocks a new capability and reuses the weights learned in the previous stage, progressing from single-view perception, to cross-view fusion and vision--language alignment, and finally to trajectory planning and joint reasoning.

\subsubsection{Stage 1: Ego-view perception pre-training.}
We first train only the ego-vehicle perception branch, namely the ego image backbone, the 3D detection head, and the vectorized map head, while the language model and the planner are disabled. The detection head predicts multi-class 3D bounding boxes together with the traffic-light state, and the map head predicts vectorized lane and boundary polylines. This stage is supervised by an ego-view detection loss and a map loss,
\begin{equation}
\mathcal{L}_{1}
= \mathcal{L}^{\text{det}}_{\text{ego}}
+ \mathcal{L}^{\text{map}}_{\text{ego}} ,
\end{equation}
where the detection loss is a set-prediction objective combining focal classification (for both object category and traffic-light state) and $\ell_1$ box regression with an auxiliary denoising term, and the map loss combines focal classification and $\ell_1$ point regression under a one-to-one plus auxiliary one-to-many assignment.

\subsubsection{Stage 2: Roadside-view perception pre-training.}
The second stage mirrors Stage~1 but trains an independent roadside branch, namely a separate roadside image backbone, roadside detection head, and roadside map head. Because the roadside camera is static and forward-facing, we use a dedicated perception range tailored to its viewpoint. The roadside branch does not estimate ego motion or traffic-light state, and is supervised against the roadside ground truth in the roadside coordinate frame,
\begin{equation}
\mathcal{L}_{2}
= \mathcal{L}^{\text{det}}_{\text{road}}
+ \mathcal{L}^{\text{map}}_{\text{road}} .
\end{equation}

\subsubsection{Stage 3: Cross-view fusion and vision--language alignment.}
The third stage loads both perception branches and jointly fine-tunes them together with the Cross-View Query Alignment and Fusion (CQAF) module and the fused detection/map heads. In parallel, the language model is activated and adapted with LoRA. This stage runs in a perception-only mode in which the planner is not yet supervised: the language model learns to align the ego, roadside, and fused detection and map queries with language through a next-token prediction loss on question--answer pairs. The objective aggregates the three perception views and the vision--language term,
\begin{equation}
\mathcal{L}_{3}
= \mathcal{L}^{\text{det}}_{\text{ego}} + \mathcal{L}^{\text{map}}_{\text{ego}}
+ \mathcal{L}^{\text{det}}_{\text{road}} + \mathcal{L}^{\text{map}}_{\text{road}}
+ \mathcal{L}^{\text{det}}_{\text{fused}} + \mathcal{L}^{\text{map}}_{\text{fused}}
+ \mathcal{L}_{\text{vlm}} .
\end{equation}

\subsubsection{Stage 4: Trajectory planning.}
The fourth stage activates the trajectory planner. The image backbones are frozen and the language model continues to be trained through LoRA. The fused detection and map semantic tokens are fed to the language model, whose waypoint hidden state conditions the planner. We support three interchangeable planning decoders: a probabilistic VAE decoder that samples a latent code from a present/future Gaussian pair, a deterministic MLP decoder that directly regresses the future waypoints, and a diffusion decoder that denoises trajectory anchors into future trajectories. In addition to the vision--language loss $\mathcal{L}_{\text{vlm}}$, this stage introduces a planning objective $\mathcal{L}_{\text{plan}}$ whose form depends on the chosen decoder. For the VAE and MLP decoders, the planning loss combines a command-conditioned trajectory regression loss, a map-boundary constraint, and a collision-avoidance constraint against the fused agent trajectories, with the VAE decoder adding a KL regularization on the latent distribution:
\begin{equation}
\mathcal{L}^{\text{VAE/MLP}}_{\text{plan}}
= \mathcal{L}^{\text{reg}}_{\text{plan}}
+ \mathcal{L}_{\text{bound}}
+ \mathcal{L}_{\text{col}}
+ \mathbb{1}_{\text{VAE}}\,\mathcal{L}_{\text{KL}} .
\end{equation}
For the diffusion decoder, the planning loss instead supervises the denoising process with a trajectory-mode classification loss, a denoised-trajectory regression loss, and a map-boundary constraint:
\begin{equation}
\mathcal{L}^{\text{diff}}_{\text{plan}}
= \mathcal{L}^{\text{cls}}_{\text{diff}}
+ \mathcal{L}^{\text{reg}}_{\text{diff}}
+ \mathcal{L}_{\text{bound}} .
\end{equation}
The overall objective is $\mathcal{L}_{4}=\mathcal{L}_{\text{vlm}}+\mathcal{L}_{\text{plan}}$.

\subsubsection{Stage 5: Joint reasoning and planning fine-tuning.}
The final stage jointly optimizes language reasoning and planning with mixed QA training, which interleaves free-form driving question-answering samples with planning samples so that the model refines its trajectories without losing its language capability. The waypoint hidden state is supervised only on samples that contain a waypoint token, while question-answering-only samples contribute solely to the language loss. The loss keeps the same composition as Stage~4, using the planning objective $\mathcal{L}_{\text{plan}}$ of the selected decoder,
\begin{equation}
\mathcal{L}_{5}
= \mathcal{L}_{\text{vlm}}
+ \mathcal{L}_{\text{plan}} .
\end{equation}

\subsubsection{Training Configuration.}
\paragraph{Random seed.}
All training runs fix the random seed to~0 and run in deterministic mode. We synchronize the seeds of Python, NumPy, PyTorch, and CUDA through the MMCV random-seed utility and enable cuDNN deterministic execution. The reported models are trained once under this setting; we do not average results over multiple seeds.

\paragraph{Optimization.}
Unless otherwise noted, Stages~1--5 share the same optimization setup. We use the AdamW optimizer with $\beta_1=0.9$, $\beta_2=0.999$, and weight decay $10^{-5}$. The learning rate follows a cosine-annealing schedule with linear warmup over $500$ iterations and a warmup ratio of $1/3$. Training is conducted in mixed precision (FP16) with dynamic loss scaling. Table~\ref{tab:training-config} summarizes the stage-specific batch size, number of epochs, base learning rate, and backbone-freezing policy used for the submitted checkpoint.

\begin{table}[!ht]
\centering
\caption{Stage-wise training hyperparameters for the submitted AURORA checkpoint. Shared optimizer and schedule settings are described in the text.}
\label{tab:training-config}
\small
\setlength{\tabcolsep}{6pt}
\begin{tabular}{@{}lcccc@{}}
\toprule
\textbf{Stage} & \textbf{Batch Size} & \textbf{Epochs} & \textbf{Base LR} & \textbf{Freeze Backbone} \\
\midrule
Stage 1 & 12 & 4 & $5{\times}10^{-5}$ & False \\
Stage 2 & 12 & 4 & $5{\times}10^{-5}$ & False \\
Stage 3 & 8 & 4 & $4{\times}10^{-5}$ & False \\
Stage 4 & 8 & 6 & $8{\times}10^{-5}$ & True \\
Stage 5 & 6 & 4 & $8{\times}10^{-5}$ & True \\
\bottomrule
\end{tabular}
\end{table}


\end{document}